\documentclass[runningheads,a4paper]{llncs}
\usepackage{llncsdoc}
\usepackage{amsfonts}
\usepackage{amssymb}
\usepackage{graphicx}
\usepackage{listings}
\usepackage{wrapfig}
\usepackage{cite}
\usepackage{atbeginend}

\let\llncssubparagraph\subparagraph
\let\subparagraph\paragraph
\usepackage[compact]{titlesec}
\let\subparagraph\llncssubparagraph

\begin{document}

\mainmatter  
\title{Event-Object Reasoning with Curated Knowledge Bases: Deriving Missing Information}
\titlerunning{Event-Object Reasoning with Curated Knowledge Bases}

\author{Chitta Baral
\and Nguyen H. Vo}

\institute{School of Computing, Informatics, and Decision Systems Engineering,\\ Arizona State University, Tempe, Arizona, USA\\
}

\maketitle

\begin{abstract}
The broader goal of our research is to formulate answers to why and how questions with respect to knowledge bases, such as AURA. One issue we face when reasoning with many available knowledge bases is that at times needed information is missing.   Examples of this include partially missing information about next sub-event, first sub-event, last sub-event, result of an event, input to an event, destination of an event, and raw material involved in an event. In many cases one can recover part of the missing knowledge  through reasoning. In this paper we give a formal definition about how such missing information can be recovered and then give an ASP implementation of it. We then discuss the implication of this with respect to answering why and how questions.
\end{abstract}

\section{Introduction}
Our work in this paper is part of two related long terms goals: answering ``How'', ``Why'' and ``What-if'' questions and reasoning with the growing body of available knowledge bases \footnote{See for example, http://linkeddata.org/.}, some of which are crowd-sourced. Although answering ``How'' and ``Why''  questions are important, so far little research has been done on them. Our starting point to address them has been to formulate answers to such questions with respect to abstract knowledge structures obtained from knowledge bases.  In particular, in the recent past we considered Event Description Graphs (EDGs) \cite{baral_answering_2012} and Event-Object Description Graphs (EODGs) \cite{baral_formulating_2013} to formulate answers to some ``How'' and ``Why'' questions with respect to the Biology knowledge base AURA \cite{chaudhri_aura:_2009}. 


Going from the abstract structures to reasoning with real knowledge bases (KBs) we noticed that the  KBs often have missing pieces of information, such as properties of an instance (of a class) or relations between two instances. For example, AURA does not encode that \textit{Eukaryotic translation} is the next event of \textit{Synthesis of RNA in eukaryote}; this may be because the two subevents of ``Protein synthesis'' were encoded independently. The missing pieces make the KB and the Description Graphs constructed from it fragmented and as a result answers obtained with respect to them are not intuitive.  Moreover, the KBs like AURA often have two or more names that refer to the same entity. To get intuitive answers they need to be resolved and merged into a single entity. Such finding of non-identical duplicates in the KB and merging them into one  is referred in the literature as entity resolution \cite{getoor2005link, brizan2006survey}.



In this paper, we start with introducing  knowledge description graphs (KDGs) as structures that can be (without much reasoning) obtained from frame based KBs such as AURA. We discuss underspecified knowledge description graphs (UDGs)  and formulate notions of reasoning with respect to these graphs to obtain certain missing information. We then present our approach of entity resolution and use it in recovering additional missing information. We give an Answer Set Programming (ASP) encoding of our formulation. We conclude with a discussion on  the use of the above in answering ``why'' and ``how'' questions.


\section{Background: Frame-based Knowledge Bases; ASP}



The KB we used in this work is based on AURA \cite{chaudhri_aura:_2009} and was described in details in \cite{baral_knowledge_2012}. AURA is a frame-based KB manually curated by biology experts; it contains a large amount of frames describing biological entities events (or processes). One important aspect of our KB is the class hierarchy. For example \footnote{Our examples are either directly from AURA, or are slightly modified from it.}: its basic class is \textit{Thing}, which has two children classes: \textit{Entity} and \textit{Event}. \textit{Entity} is the ancestor of all classes of biological entities;  \textit{Event}, of biologicalogy events. For instance, \textit{Spatial entity}, \textit{Eukaryote}, \textit{Nucleus} and \textit{mRNA} are descendants of \textit{Entity}, while ``Eukaryotic translation", ``Eukaryotic transcription" are descendant of \textit{Event}.

Our KB is a set of facts of the form ``has($A$, slot\_name, $B$)'' where $A$ and $B$ are either classes or instances (of classes), \textit{slot\_name} is the name of the relation between $A$ and $B$ such as \textit{instance\_of}, \textit{raw\_material} or \textit{results}. 
The statement ``\textit{eukaryotic translation} is based on \textit{mRNA}'' is represented in our KB as follows. 
\begin{lstlisting}
has(euka_transl4191, instance_of, event).
has(euka_transl4191, instance_of, eukaryotic_translation).
has(euka_transl4191, base, mrna4642).
has(mrna4642, instance_of, mrna).
\end{lstlisting}
This snippet reads as ``\textit{eukaryotic\_translation4191} is an instance of class \textit{event} and an instance of class \textit{eukaryotic\_translation}. \textit{eukaryotic\_translation4191} is based on \textit{mrna4642}, which is an instance of \textit{mrna}''.
  
For the declarative implementation of our formulations, we use ASP \cite{gelfond_stable_1988}. That allows us to use our earlier work  \cite{baral_knowledge_2012} on using ASP to reason with frame-based knowledge bases. ASP's  strong theoretical foundation \cite{baral_knowledge_2003} and its default negation and recursion are useful in our encoding and in proving results about them.

\section{Knowledge Description Graphs}
 An {\bf Underspecified Knowledge Description Graph (UDG)} is a structure to represent the facts about instances and classes of events, entities and relationships between them. An UDG is constructed from knowledge bases such as AURA. Formal definition of the UDGs is given in the following.
\begin{definition}
\label{def:kdg}
An UDG is a directed graph with one type of node and five types of directed edges: compositional edges, ordering edges, class edges, locational edges and participant edges. Each node represents an instance (of a class) or a class in our KBs. 
\end{definition}

\begin{table}[h]
	\centering
	\begin{tabular}{lll}
		\hline\hline
 	Edge type & 	Relation(s)  \\
		\hline
	locational &	happenings  \\
	class & instance-of, super-class \\
	compositional &	subevent, first-subevent, has-part, has-region, has-basic-structural-unit  \\
	ordering &	next-event, enables, causes, prevents, inhibits \\
	participant & raw-materials, result, agent, destination, instrument, origin, site  \\
	\hline\hline
	\end{tabular}
	\caption{Types of edges in an UDG. An edge of relation $Y$ from a node $X$ to $Z$ represents $X[Y] = Z$, meaning the slot $Y$ of the entity $X$ has value $Z$.}
	\label{table:Types_of_edges_UDG}
\end{table}

We used the slot names in KM \cite{clark_km:_2004} and AURA as a guide to categorize four types of edges  (Table \ref{table:Types_of_edges_UDG}). 

\begin{figure}
\centering
\includegraphics[width=0.8\textwidth]{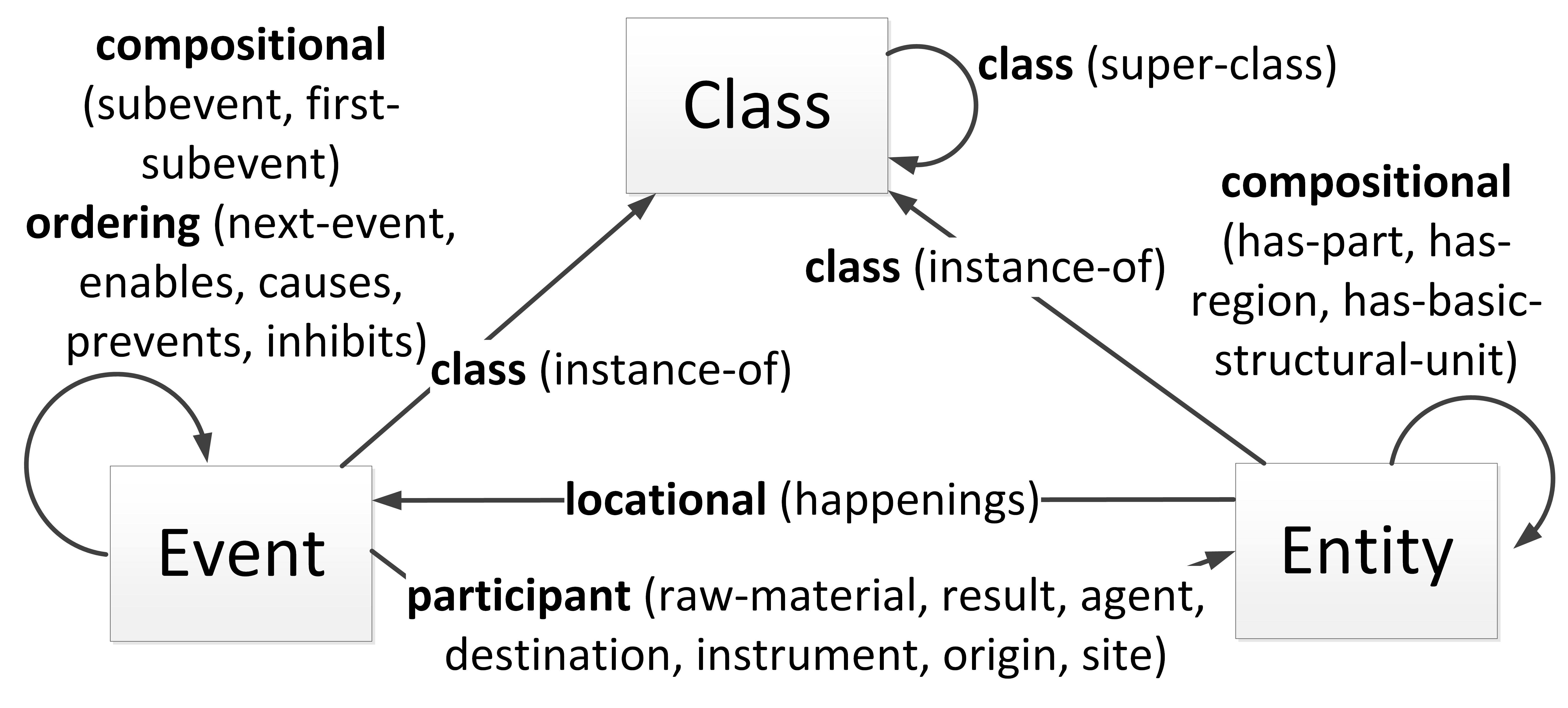}
\caption{Types of edges in a KDG }
\label{fig:Types_of_edges}
\vspace{4pt}
\end{figure}


A {\bf Knowledge Description Graph (KDG)} (a slight generalization of EODGs in \cite{baral_formulating_2013}) 
is constructed from an UDG. A node in a KDG represents either an instance of a biological entity, an instance of a biological process, or a class of biological entity/event. The KDG structure allows us to answer ``How'' and ``Why'' questions. More formally, a KDG is defined as follows.

\begin{definition}
\label{def:KDG}
A KDG is a directed graph with: (i) three types of nodes: event nodes, entity nodes, and class nodes; and (ii) five types of directed edges: compositional edges, class edges, ordering edges, locational edges and participant edges. A KDG has the property that there are no directed cycles within any combination of compositional, locational and participant edges.
\end{definition}

Figure \ref{fig:Types_of_edges} shows the types of edges in a KDG and the corresponding sources and destinations of the edges. Edges in a KDG are from the edges of the UDG, with additional type constraints of  the source and destination nodes. For example, ordering edges must be from events to events; compositional edges are from events to events or from entity to entity, depending on their specific relations.

Since UDGs and KDGs can be huge, we usually work on their smaller subgraphs that are rooted at an entity or an event.  They are defined as follows.

\vspace{-5pt}
\begin{definition}
\label{def:rooted_kdg}
Let $Z$ be a node in a KDG $G$. The Knowledge Description Graph (KDG) rooted at $Z$ is the subgraph of $G$ composed of:  (1)  The set $N$ of all the nodes of $G$ that are accessible from $Z$ through compositional edges, class edges, locational edges or participant edges; and  (2) All the edges of $G$ connecting two nodes in $N$.  We denote the KDG rooted at $Z$ as $KDG(Z)$ or the KDG of $Z$.
\end{definition}
\vspace{-5pt}

\begin{figure}
\centering
\includegraphics[width=1\textwidth]{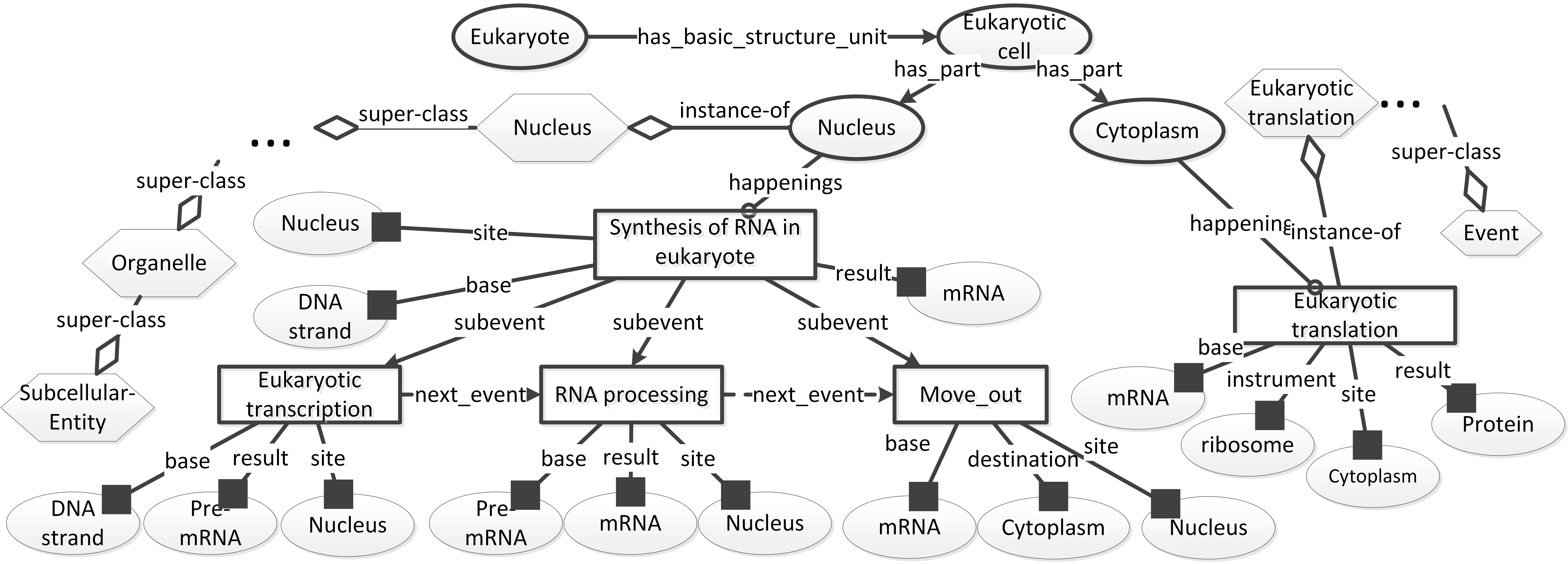}
\caption{A KDG rooted at the entity \textit{Eukaryote}. Event, entity, and class nodes are respectively depicted by rectangles, circles and hexagons. Compositional edges are represented by solid-line arrows; ordering edges by dashed-line arrows; participant edges by lines with a black-square pointed to the entity node; class edges by diamond head arrows and locational edges by lines with a circle pointed to the event node.}
\label{fig:KDG_eukaryotes}
\end{figure}

The UDG rooted at $Z$, denoted as $UDG(Z)$, is defined similarly. Figure ~\ref{fig:KDG_eukaryotes} shows an example of a KDG rooted at \textit{Eukaryote} where every other nodes can be reached from \textit{Eukaryote} through edges with solid lines (compositional edges, class edges, locational edges or participant edges).
 
%
%

\section{Reasoning about Missing Info. in UDGs and KDGs}
\label{sec:methods}
In this section, we discuss about missing information in the UDGs and the KDGs and how we can recover some of it 
through reasoning.



\subsection{Event, Next Event, First Sub-event and Last Sub-event}

\label{sec:event_first_subevent}
One can directly obtain event names by looking at facts of the form ``has(E, instance\_of, event)'' in the KB; and concluding E in it as an event. However, for some events such facts may be missing. In that case, we may be able to get the fact from the UDG's edges and the edge constraints of the KDGs (Figure \ref{fig:Types_of_edges}). More formally:
\begin{definition}
	Let $E$ be a node in the $UDG(Z)$. $E$ is an event if there is
	(i) a participant edge or an ordering edge from $E$;
	(ii) a locational edge or an ordering edge to $E$;
	(iii) a compositional edge (of subevent or first subevent relation) from/to $E$; or
	(iv) a path of class edges from $E$ to the class \textit{event}.
	\label{def:event}
\end{definition}
Based on Definition \ref{def:event}, we can get that \textit{photosynthesis} is an event because it has compositional edges (of subevent relation) to \textit{light reaction} and \textit{calvin cycle}.

Next-event, first subevent and last subevent are amongst the most important properties in describing events. However, they are not always directly available in our KB. Fortunately, in many cases, we can recover them from other properties.
\vspace{-4pt}
\begin{definition}
	Let $E$ and $E'$ be two events in the $KDG(Z)$.
	Event $E'$ is the next event of $E$ if $E$ enables, causes, prevents or inhibits $E'$.
	\label{def:next_event}
\end{definition}
\vspace{-4pt}
In other words, $E'$ is the next event of $E$ if there is an ordering edge from $E$ to $E'$. 

\vspace{-4pt}
\begin{definition}
	Let $S$ be the set of subevents of an event $X$ in the $KDG(Z)$. Event $E$ in $S$ is the first subevent of $X$ if there exists no other event $E'$ in $S$ such that $E$ is the next event of $E'$.
	Similarly, event $E$ in $S$ is the last subevent of $X$ if there exists no other event $E'$ in $S$ such that $E'$ is the next event of $E$.
	\label{def:first_last_event}
\end{definition}
\vspace{-4pt}

Here we assume that $S$ was properly encoded in that there is only one chain of subevents in $S$. In our KB, \textit{light reaction} and \textit{calvin cycle} are two subevents of \textit{photosynthesis} and \textit{light reaction} enables \textit{calvin cycle}. But their orders are not defined. However, using Definition \ref{def:next_event}  and \ref{def:first_last_event}, we can identify that: \textit{calvin cycle} is the next event of \textit{light reaction}; \textit{light reaction} is the first subevent of \textit{photosynthesis}; and \textit{calvin cycle} is the last subevent. 

\subsection{Input/Output of Events}
\textbf{Two types of events:}
In our KB there are two types of events: transport events and operational events. In a transport event, there is only a change in the locations; the input location and output location are different from each other while the input entity and output entity are the same. All other events are operational events. In an operational event, there is usually no change in its location. We differentiate two types of events by their ancestor classes; transport events are descendants of the classes \textit{move\_through}, \textit{move\_into} and \textit{move\_out\_of}.


\textbf{Input, Output, Input Location, Output Location: }
To reason about the KDG, we need the input and output of each event as well as the input location and the output location, which are not always available. In the following, we show how to use various event's relations - such as raw-material, destination, location and others - to create four new relations (IO relations): input, output, input-location and output-location. After that, we propose rules to complete the KDG's IO relations. 

We created the IO relations of an event based on specific relations as shown in Table \ref{table:IO_props_of_events}. The meaning of relation ``base'' from AURA depends on the context. For transport events, it is for input-location; for operational events, it is for input. 

\begin{table}[h]
	\centering
	\begin{tabular}{lll}
		\hline\hline
 	Event type & 	IO relation type        &      Relation(s)  \\
		\hline
	Transport event &	input    &   object  \\
	Transport event &	output      &   object \\
	Transport event &	input-location &   base, origin \\
	Transport event &	output-location    &   destination  \\
	Operational event &	input    &   object, base, raw-material  \\
	Operational event &	output      &   result \\
	Operational event &	input-location &   site \\
	Operational event &	output-location &  destination \footnote{The destination slot of an operational event is usually not defined in the AURA}   \\
	\hline\hline
	\end{tabular}
	\caption{The IO properties of events and their corresponding relations.}
	\label{table:IO_props_of_events}
	\vspace{-10pt}
\end{table}

\textbf{Completing Missing Information of Input, Output, Input Location, Output Location:} We can obtain missing IO properties of an event from its subevent(s). For instance, an input of the first subevent of $E$ is also an input of $E$. 
 

\vspace{-4pt}		
\begin{definition}
	\label{def:IO_from_subevents}
	Let $FSE$ and $LSE$ respectively be the first subevent and last subevent of event $E$ in $KDG(Z)$. 
	
	Let $InputRelation$ be the input relation, input-location relation or one of their corresponding relations (Table \ref{table:IO_props_of_events}). If $InputRelation$ is a relation from $FSE$ to $X$ then $InputRelation$ is also a relation from $E$ to $X$.
	
	Let $OutputRelation$ be the output relation and output-location relation or one of their corresponding relations. If $OutputRelation$ is a relation from $LSE$ to $X$ then $OutputRelation$ is also a relation from $E$ to $X$.
\end{definition}
\vspace{-4pt}

In our KB, \textit{photosynthesis} has two subevents: \textit{light reaction} and \textit{calvin cycle}, the next event of \textit{light reaction}. \textit{Sunlight} is the raw-material of the \textit{light reaction}, \textit{sugar} is the result of \textit{calvin cycle}. Using Definition \ref{def:IO_from_subevents}, we have that \textit{sunlight} is the raw-material of \textit{photosynthesis} and \textit{sugar} is its result. Moreover, we also have: \textit{sunlight} is the input of \textit{light reaction} as well as \textit{photosynthesis}; \textit{sugar} is the output of both \textit{calvin cycle} and \textit{photosynthesis}.

Similarly, the output location of an operational event is often not defined in the KB but we can use input location as the default value for output location.

\begin{definition}
	\label{def:default_output_loc}
	Let $E$ be an event in $KDG(Z)$, $E$'s input location is also the output location if $E$'s output location has not been specified.
\end{definition}

\begin{figure}[h]
    \vspace{4pt}
	\begin{center}
		\includegraphics[width=0.75\textwidth]{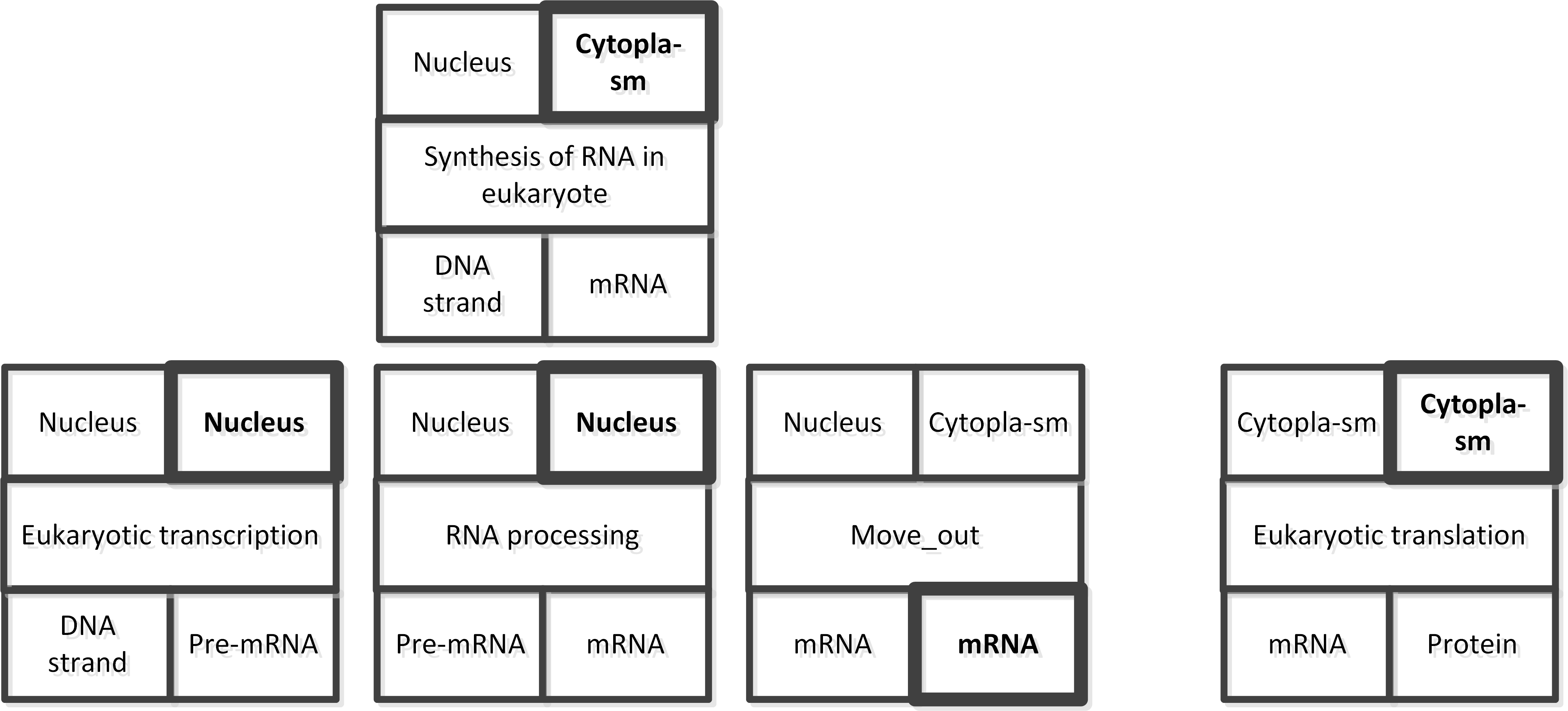}
	\end{center}
	\caption{\textit{The IO properties of events in Fig ~\ref{fig:KDG_eukaryotes}. The five blocks contain IO properties of events: \textit{Synthesis of RNA in eukaryote}, \textit{Eukaryotic translation}, \textit{Move\_out}, and \textit{Eukaryotic transcription}. The middle rectangle of each block contains the event name. The top rectangles are for input and output locations; the bottom rectangles are for input and output. The properties in bold were recovered using Definitions \ref{def:IO_from_subevents} and  \ref{def:default_output_loc}}}
	\label{fig:IO_properties_of_events}
\end{figure}

Figure \ref{fig:IO_properties_of_events} shows the IO properties of events in Fig ~\ref{fig:KDG_eukaryotes}. The properties in bold are the ones that were recovered using Definitions \ref{def:IO_from_subevents} and  \ref{def:default_output_loc}.

\subsection{Main Class of an Instance}
In our KB, one instance can belong to many classes. For example, \textit{dna\_strand19497} - the input of \textit{Eukaryotic transcription} - is an instance of \textit{dna\_strand, dna\_sequence, nucleic\_acid} and \textit{polymer} \footnote{For the sake of simplicity, in the previous figures and descriptions, we usually referenced the entities and events by their ``main'' class(es) and not by the instances' names although our KB and our implementation works on instances' names.}. However, to reason about the equality between instances, we need the ``main'' class(es), the most specific class(es) of that instance. Our formal definition of ``main'' class is given below.

\begin{definition}{}
	\label{def:Main_class}
 Let $E$ be an instance in $KDG(Z)$. 
$ClassB$ is a main class of instance $E$ if 
(1) it is a class of $E$ and 
(2) it is not the case that there is a 
$ClassA$ which is a class of $E$ and (a) $ClassB$ is ancestor of $ClassA$ or  (b) $ClassB$ is a general class but $ClassA$ is not; where general classes in our KB are \textit{thing}, \textit{event}, \textit{entity}, \textit{spatial\_entity}, \textit{tangible\_entity}, and \textit{chemical\_entity}. 
\end{definition}


The main classes of \textit{dna\_strand19497}, according to the Definition \ref{def:Main_class}, are \textit{ dna\_strand} and \textit{dna\_sequence}; the other classes of \textit{dna\_strand19497} are ancestors of those two. 

\section{Entity Resolution}
\label{sec:matching_instances}
In the KBs such as AURA, the curation was done in many sessions and probably by many people. (Same is true with respect to many other KBs; especially the ones that are developed using crowd-sourcing.) The results are, in many cases, (i) two different instance names were used when they are probably the same instance; and (ii) parts of some biological process were encoded as independent events. For example: the input of \textit{Eukaryotic translation} (Figure \ref{fig:KDG_eukaryotes}) is \textit{mrna4642} whereas the output of \textit{Move\_out} is \textit{mrna22911}; \textit{Synthesis of RNA in eukaryote} and \textit{Eukaryotic translation} should be subevents of ``Synthesis of protein in eukaryote'' but they are encoded as two separate events.

In this section, we propose methods to solve the first problem. These methods are then used to solve the second problem in the next section.
In order to compare two instances in a KB, we define a match relation. Generally speaking, instance $A$ can match with instance $B$ if $A$ can be safely used in a context where a term of $B$ is expected. We defined matching relation with many confidence levels for greater flexibility in future works.

\begin{definition}
	\label{def:Matching_instance}
	Let $A$ and $B$ be two instances in $KDG(Z)$. Let $ClassA$ and $ClassB$ be main classes of $A$ and $B$ respectively.
	\begin{enumerate}
	\item $A$ matches with $B$ with high confidence if one of the following is true
		\begin{enumerate}
			\item $A$ = $B$ ($A$ and $B$ are the same instance)
			\item $A$ is cloned from $B$ (Shortcut in AURA to specify that $A$ has all the properties of $B$)
			\item $ClassA$ is an ancestor of $ClassB$.
		\end{enumerate}
	\item $A$ matches with $B$ with medium confidence if $A$ and $B$ are both cloned from an instance $C$.
	\item $A$ matches with $B$ with low confidence if $ClassA$ = $ClassB$ ($A$ and $B$ are instances of the same main class).
	\item $A$ matches with $B$ with confidence $min(Conf_1, Conf_2)$  if 
		\begin{enumerate}
			\item $A$ matches with $C$ with confidence $Conf_1$ and
			\item $C$ matches with $B$ with confidence $Conf_2$
		\end{enumerate}
	\item Otherwise, $A$ does not match with $B$.
	\end{enumerate}
\end{definition} 

Using Def.\ref{def:Matching_instance}, we can match \textit{mrna4642} - the input of \textit{Eukaryotic translation} - with \textit{mrna22911} - the output of \textit{Move\_out}, because both have \textit{mrna} as the main class.

While Def.\ref{def:Matching_instance} can match all the input and output in our aforementioned example, it is not sufficient for matching location. For example, we can not match an instance of \textit{cytoplasm} to an instance of \textit{cytosol}. However when we say \textit{Event $A$ occurs in cytosol}, we can understand that \textit{Event $A$ occurs in cytoplasm}. To overcome this shortcoming, we define the relation \textit{Spatially match} as follows.

\begin{definition}
	\label{def:Location_instance}
	Instance $A$ in $KDG(Z)$ is a location instance if the class $ClassA$ of $A$ is a descendant of the class $spatial\_entity$.
\end{definition}
		
\begin{definition}
	\label{def:Spatially_Matching_instance}
	Let $A$ and $B$ be two location instances in $KDG(Z)$. Let $ClassA$ and $ClassB$ be main classes of $A$ and $B$ respectively.
	\begin{enumerate}
		\item Location $A$ spatially matches with location $B$ with confidence $Conf$ if instance $A$ matches with instance $B$ with confidence $Conf$.
		\item Location $A$ spatially matches with location $B$ with high confidence if one of the following is true:
		\begin{enumerate}
			\item $B$ is inside $A$ (the relation \textit{inside} is encoded in our KB by slot name \textit{is\_inside}).
			\item $B$ is part of $A$ (the relation \textit{``part of''} is encoded in our KB by slot name \textit{part\_of}).
		\end{enumerate}
		\item Location $A$ spatially matches with location $B$ with confidence $min(Conf_1, Conf_2)$ if
		\begin{enumerate}
			\item $A$ spatially matches with location $C$ with confidence $Conf_1$, and
			\item $C$ spatially matches with $B$ with confidence $Conf_2$.
		\end{enumerate}
	\end{enumerate}
\end{definition} 
Suppose that  in our KB we have: \textit{cytosol234 and cytosol987 all are instances of cytosol; cytoplasm322 is an instance of cytoplasm and cytosol987 is inside cytoplasm322}. We can then conclude: \textit{cytosol234 and cytosol987 match with each other with low confidence, according to Def.\ref{def:Matching_instance}.3; cytoplasm322 spatially matches with cytosol987 with high confidence (Def.\ref{def:Spatially_Matching_instance}.2.a); cytosol987 spatially matches with cytosol234 with low confidence (Def.\ref{def:Spatially_Matching_instance}.1); and cytoplasm322 spatially matches with cytosol234 with low confidence (Def.\ref{def:Spatially_Matching_instance}.3) }.

\section{Finding the Possible Next Events}
\label{sec:possible_next_events}
In this section, we demonstrate the usefulness of matching instances (Definition \ref{def:Matching_instance} and \ref{def:Spatially_Matching_instance}) in finding the possible next event(s) of a given event. 
While in simple cases (Section \ref{sec:event_first_subevent}), we can find the next event $E$' of an event $E$ by using Definition \ref{def:next_event}, there are still cases where there exists no ordering edges from $E$ to $E'$. For examples, \textit{Alteration of mrna ends} and \textit{RNA splicing} are two subevents of \textit{RNA processing} but no other relation between them was defined. However, they all occur in \textit{nucleus16421} and \textit{Alteration of mrna ends}'s output, \textit{pre\_mrna7690}, matches with \textit{RNA splicing}'s input, \textit{rna8697}. This information hints us that \textit{RNA splicing} is \textit{Alteration of mrna ends}'s next event. 

Following this intuition, our approach for finding the next event is that $E'$ is the next event of $E$ if the output of $E$ matches the input of $E'$ and output location of $E$ matches the input location of $E'$. In the example in Figure \ref{fig:KDG_eukaryotes}, this assumption holds in all three events: \textit{Eukaryotic transcription}, \textit{RNA processing} and \textit{Move\_out}; all of which are already defined in our KB as consequent events. This assumption also suggests that \textit{Eukaryotic translation} can be the next event of either \textit{Synthesis of RNA in eukaryote} or \textit{Move\_out}. Armed with Definition \ref{def:Matching_instance} and \ref{def:Spatially_Matching_instance}, we define the following join relation.

\begin{definition}
	Let $A$ and $B$ be two events in $KDG(Z)$.
	Event $A$ joins to event $B$ if all of the following conditions are true:
	\label{def:joinable_event}
	\begin{enumerate}
		\item The output of $A$ matches with the input of $B$ or vice versa.
		\item The output location of $A$ spatially matches with the input location of $B$ or vice versa.
	\end{enumerate}
\end{definition} 

Applying this definition, we have: \textit{Alteration of mrna ends} joins to \textit{RNA splicing}, \textit{Eukaryotic transcription} joins to \textit{RNA processing}; \textit{RNA processing} joins to \textit{Move\_out}; and both \textit{Synthesis of RNA in eukaryote} and \textit{Move\_out} are joined from \textit{Eukaryotic translation}.
Since we want \textit{Eukaryotic translation} to be a possible next event of \textit{Synthesis of RNA in eukaryote} instead of its subevent \textit{Move\_out}, we define the possible next event as follows.

\begin{definition}
	\label{def:Possible_next_event}
	Let $A$ and $B$ be two events in $KDG(Z)$ where $A$ joins to $B$. $B$ is a possible next event of $A$ if none of the following conditions is true:
	\begin{enumerate}
		\item $A$ joins to $AncestorB$ where $AncestorB$ is the ancestor event of $B$ (in other words, there is a non-empty path of subevent relation from $AncestorB$ to $B$).
		\item Ancestor event $AncestorA$ of $A$ joins to $B$.
		\item $A$ is an ancestor event of $B$.
		\item $B$ is an ancestor event of $A$.
		\item $A$ and $B$ have the same ancestor event.
	\end{enumerate}
\end{definition}


In our example, condition \ref{def:Possible_next_event}.2 gives us that \textit{Eukaryotic translation} is not the possible next event of \textit{Move\_out} while \ref{def:Possible_next_event} concludes that \textit{Eukaryotic translation} is the possible next event of \textit{Synthesis of RNA in eukaryote}. We assume that an event and its subevents are put in our KB as a whole, so the \textit{next\_event} relations between them are well defined. Thus conditions \ref{def:Possible_next_event}.3-5 take those relations out of consideration.

When we have a path of possible next events, we can create an event $SE$, which is the super event of all events in the path, and add suitable \textit{next\_event} or \textit{subevent} relations. This new event would link the events that were mistakenly encoded as independent events (that we mentioned earlier).


\section{ASP Encodings}

In this section we give ASP encodings of our formulations in the previous sections. 

%


\textbf{Encoding the Entities and Events:} 
Rules t1-t2 in the following state that an instance $X$ is an event or an entity if and only if it is the instance of $event$ class or $entity$ class respectively. Rules t3-t4 identify $E$ as an event if there is an ordering edge to $E$ (Definition \ref{def:event}.ii). Rule t5 encodes Definition \ref{def:event}.iv. ``has(X, ancestorclass, Y)'' denotes the transitive closure of ``has(M, superclass, N)'' and is encoded the standard way (rules t6-t7). The rest of Definition \ref{def:event} are encoded similarly in rules t8-t21. 

\begin{lstlisting}
t1: event(X):-has(X,instance_of,event).
t2: entity(X):-has(X,instance_of,entity).
t3: ordering_edge(next_event; enables; causes; prevents; inhibits).
t4: has(E, instance_of, event) :- has(X, S, E), ordering_edge(S). 
t5: has(E, instance_of, event) :- has(X, instance_of, ClassY), has(ClassY, ancestorclass, event).

\end{lstlisting}

%
%
%

\textbf{Finding Next Events, First Subevents and Last Subevents:} 
Rules e1-e2 find the next events (Definition \ref{def:next_event}) and rules e3-e6 find the first subevents and the last subevents (Definition \ref{def:first_last_event}).
\begin{lstlisting}
e1: predicates(ordering_edge, enables; causes; prevents; inhibits).
e2: has(E1, next_event, E2) :- has(E1, Predicate, E2), predicates(ordering_edge, Predicate).
e3: not_fse(Z, E) :- has(Z, subevent, E), has(Z, subevent, E2), E2 != E, has(E2, next_event, E).
e4: not_lse(Z, E) :- has(Z, subevent, E), has(Z, subevent, E2), E2 != E, has(E, next_event, E2).
e5: has(Z, first_subevent, E) :- has(Z, subevent, E), not not_fse(Z, E).
e6: has(Z, last_subevent, E) :- has(Z, subevent, E), not not_lse(Z, E).
\end{lstlisting}

\textbf{Encoding Transport Events and Operational Events:} 
$t\_event(E)$ or $o\_event(E)$ is used to indicate a transport event or an operational event, respectively.

\begin{lstlisting}
ev1: predicates(t_event, move_through; move_into; move_out_of).
ev2: t_event(E) :- has(E, instance_of, Transport_class), predicates(t_event, Transport_class), event(E).
ev3: o_event(E) :- event(E), not t_event(E).
\end{lstlisting}

\textbf{Encoding the Inputs and Outputs of Operational Events:}
We denote the input/output/input location/output location of an event by $input$, $output$, $input\_loc$ and $output\_loc$ respectively. Rules i1-i5 get the IOs of operational events. IOs of transport events are encoded similarly (rules i6-i10).

\begin{lstlisting}
i1:input(E,A):-has(E,object,A),o_event(E).
i2:input(E,A):-has(E,base,A),o_event(E).
i3:input(E,A):-has(E,raw_material,A),o_event(E).
i4:output(E,A):-has(E,result,A),o_event(E).
i5:input_loc(E,A):-has(E,site,A),o_event(E).
\end{lstlisting}

\textbf{Getting the Missing Inputs and Outputs:}
Rule i11 gets the input of an event from its first subevent (Definition \ref{def:IO_from_subevents}.1). Rule i12 gets the \textit{object} property of a transport event from its first subevent. Other rules to get the input location, output and output location as well as other properties, such as raw-material, result, are encoded in a similar way (rules i13-i24). Rule i25 gets the default output location of an event(Definition \ref{def:default_output_loc}).

\begin{lstlisting}
i11: has(E, input, A) :-  has(SE, input, A), has(E, first_subevent, SE).
i12: has(E, object, A) :- has(SE, object, A), has(E, first_subevent, SE), transport_event(E).
i25: has(E, output_location, A) :- not has(E, output_location, A2), has(E, input_location, A),  entity(A2), event(E), A2 != A. 
\end{lstlisting}

\textbf{Encoding the Main Class(es) of an Instance:}
$ClassA$ is a main class of instance $A$ if $ClassA$ is one of $A$'s classes and we do not have $not\_main\_class(A, ClassA)$ (which mean $ClassA$ is not the main class of $A$).

\begin{lstlisting}
m1: general_class(thing; event; entity; spatial_entity; tangible_entity; chemical_entity).
m2: not_main_class(A, ClassB) :- has(A, instance_of, ClassA), has(A, instance_of, ClassB), has(ClassA, ancestorclass, ClassB). 
m3: not_main_class(A, ClassB) :- has(A, instance_of, ClassA)), has(A, instance_of, ClassB), general_class(ClassB), not general_class(ClassA).
m4: main_class(A, ClassA) :- has_class(A, ClassA), not not_main_class(A, ClassA).
\end{lstlisting}

\textbf{Encoding Instance Matching:}
We use predicate $match\_with(A, B, Confidence)$ to represent \textit{match with} relation (Definition \ref{def:Matching_instance}) from instance $A$ to $B$; $Confidence$ can be either \textit{low, medium} or \textit{high}. Rule ma1 encodes the sub-case \ref{def:Matching_instance}.1.a of Definition \ref{def:Matching_instance}. The last rule is for Definition \ref{def:Matching_instance}.4, matching $A$ to $B$ transitively through $C$. $lowest\_confidence(Conf1, Conf2, Conf)$ means $Conf$ is the lowest confidence in $Conf1$ and $Conf2$ (Rules lc1-lc7). Rules for other cases of Definition \ref{def:Matching_instance} are skipped (rules ma2-ma5); locational instance matching is encoded in a similar way (rules sma1-sma4).
\begin{lstlisting}
ma1:  match_with(A,B,high) :- main_class(A,ClassA), main_class(B,ClassB), A==B.
ma6:  match_with(A,B,Conf) :- match_with(A,C,Conf1), match_with(C,B,Conf2), A!=B, A!=C, B!=C, lowest_confidence(Conf1,Conf2,Conf).
\end{lstlisting} 

\textbf{Encoding Possible-next-event Relation:}
In this section, we show how Definition \ref{def:Possible_next_event} is encoded. We use $has(A, tc\_subevent, B)$
to represent transitive closure of \textit{sub event} relation between $A$ and $B$ (encoded by $has(A, subevent, B)$), which is defined in the standard way (rules tcsub1-tcsub3). We also use $\_join(A, B)$ to encode that $A$ joins to $B$ according to Definition \ref{def:joinable_event} (rules j1-j3). The two rules below is corresponding to the sub-case \ref{def:Possible_next_event}.1. Other cases are skipped (rules n2-n5).
\begin{lstlisting}
n1: _notNextEvent(A, B) :- _join(A, SuperB), _join(A, B), has(SuperB, tc_subevent, B).
n6: possible_next_event(A,B) :- _join(A, B), not _notNextEvent(A, B).
\end{lstlisting}
\textbf{Correctness of the ASP Rules:}
\begin{definition}
	The ASP program $\Pi_{Z}$ is the answer set program consisting of the facts of the form ``$has(X,S,V)$'' that are generated from all the nodes and edges of $KDG(Z)$ in the following way:
	\begin{enumerate}
	\item For each node $N$, generate ``$has(N,instance\_of, event)$'' if $N$ is event node, ``$has(N,instance\_of, entity)$ if $N$ is entity nodes.
	\item For each edge of relation $R$ (Table \ref{table:Types_of_edges_UDG}) from $E1$ to $E2$, generate `$has(E1,R,E2)$''.
	\end{enumerate} 
\end{definition}

\begin{definition}
The ASP program $\Pi$ is the answer set program consisting of the following rules:
t1 to t14 for events and entities,
e1 to e6 for next events, first subevents and last events, 
ev1 to ev3 for two types of events, 
i1 to 25 for inputs, outputs of events,  
m1 to m4 for main class(es), 
lc1 to lc7 for the lowest confidence,
ma1 to ma6 for match relation,
sma1 to sma4 for spatially match relation,
tcsub1 to tcsub2 for transitive closure of subevents, 
j1 to j3 for joined events, and
n1 to n6 for possible next events.
\end{definition}

\noindent
\textbf{Proposition 1: } $E$ is the last subevent of $X$ in $KDG(Z)$ iff \\
	$
		\Pi_Z \cup \Pi \models has(Z, last\_subevent, E)
	$
\\ \noindent
\textbf{Proposition 2: } $A$ is the main class of $E$ in $KDG(Z)$ iff \\
	$
		\Pi_Z \cup \Pi \models main\_class(E, A)
	$
\\ \noindent
\textbf{Proposition 3: } 	Let $A$ and $B$ be two instances in $KDG(Z)$. $A$ matches with $B$ with the confidence level $Conf$ iff
	$
		\Pi_Z \cup \Pi \models match\_with(A, B, Conf)
	$
\\ \noindent
\textbf{Proposition 4: } 	Let $A$ and $B$ be two events in in $KDG(Z)$. $A$ is a possible next event of $B$ iff
	$
		\Pi_Z \cup \Pi \models possible\_next\_event(A,B)
	$

\section{Discussion: Answering ``How'' and ``Why'' Questions}

In Section \ref{sec:methods}, we showed how to recover missing information using properties of KDG's structure. Completing this information not only allows us to improve the KB that was used to construct the KDG, but also make it possible to reason about large curated KB using KDG. 
In Section \ref{sec:matching_instances} and \ref{sec:possible_next_events}, we also solved an important step in bringing the KB's usage out of small examples: we proposed the methods to compare instances and demonstrated their power in finding possible next events.

Those efforts have enabled us to answer deep reasoning questions, such as ``How'' and ``Why'' questions. We give examples of a few of them in the following. Details about answering them are explained in another work of ours \cite{baral_formulating_2013}.

\begin{enumerate}
\item The answer of ``How does X occur?'' is simply a structure that basically contains $KDG(X)$ and all the nodes connected to/from $X$ through ordering edges.
\item The answer of ``How does X produce Y?'' is similar to ``How does X occur?'' but $X$ must produce $Y$.
\item The answer of ``How are X and Y related?'' is a simplified structure of $KDG(Z)$ that contains: two paths of component edges to  $X$ and $Y$ from their lowest common ancestor and all paths of ordering edges linking two nodes in those two paths.
\item Similarly, the answer of ``Why X is important to Y?'' is the answer of ``How are X and Y related?'' plus the path on ``important'' links which explains why $X$ is important to $Y$. An ``important'' link from $A$ to $B$ is defined in $KDG$ to indicates that $A$ is important to $B$.
\item Other questions that KDG can answer includes ``How does X participate in process Y?'', ``How does X do Y?'', ``Why does X produce Y?'' and others.

\end{enumerate}

\section{Conclusion}

In this paper we have shown how to derive certain missing information from large knowledge bases. Often such knowledge bases are created by multiple people; sometimes even through crowd-sourcing. This often leads to some information being not explicitly stated, even though the knowledge base contains clues to derive that information. In our larger quest to formulate answers to ``why'' and ``how'' questions, we focused on the frame based knowledge base AURA, noticed several such omissions, and using those as examples, developed several general formulations regarding missing knowledge about events.  We also gave an ASP implementation of our formulations and used them in answering ``why'' and ``how'' questions. We briefly discussed some of those question types and how their answer can be obtained from Knowledge Description Graphs (KDGs). Thus, by being able to obtain missing information and enriching the original  KDGs one can obtain more accurate and intuitive answers to the various `why'' and ``how'' questions.

One of our formulations was about entity resolution where we resolve multiple entities that may have different names but may refer to the same entity. Our method is different from other methods in the literature \cite{getoor2005link, brizan2006survey}. Since each entity resolution method heavily relies on the properties of the database it is working on, and no other system we know of is about AURA or similar event centered knowledge bases we were unable to directly compare our method with the others. 

Our approach to use rules (albeit ASP rules)  to derive missing information is analogous to use of rules in data cleaning  and in improving data quality \cite{herzog2007data,rahm2000data,fan2009conditional}. However those works do not focus on issues that we discussed in this paper.

\bibliographystyle{splncs}
\bibliography{lpnmr13}

\begin{thebibliography}{10}

\bibitem{baral_answering_2012}
Baral, C., Vo, N.H., Liang, S.:
\newblock Answering why and how questions with respect to a frame-based
  knowledge base: a preliminary report.
\newblock Technical Communications of the 28th International Conference on
  Logic Programming (ICLP'12) \textbf{17} (2012)  26--36

\bibitem{baral_formulating_2013}
Baral, C., Vo, N.:
\newblock Formulating question answering with respect to event-object
  description graphs.
\newblock Unpublished paper submitted to a conference (2013)

\bibitem{chaudhri_aura:_2009}
Chaudhri, V.K., Clark, P.E., Mishra, S., Pacheco, J., Spaulding, A., Tien, J.:
\newblock {AURA:} capturing knowledge and answering questions on science
  textbooks.
\newblock Technical report, {SRI} International (2009)

\bibitem{getoor2005link}
Getoor, L., Diehl, C.P.:
\newblock Link mining: a survey.
\newblock ACM SIGKDD Explorations Newsletter \textbf{7}(2) (2005)  3--12

\bibitem{brizan2006survey}
Brizan, D.G., Tansel, A.U.:
\newblock A survey of entity resolution and record linkage methodologies.
\newblock Communications of the IIMA \textbf{6}(3) (2006)  41--50

\bibitem{baral_knowledge_2012}
Baral, C., Liang, S.:
\newblock From knowledge represented in frame-based languages to declarative
  representation and reasoning via {ASP}.
\newblock 13th International Conference on Principles of Knowledge
  Representation and Reasoning (2012)

\bibitem{gelfond_stable_1988}
Gelfond, M., Lifschitz, V.:
\newblock The stable model semantics for logic programming.
\newblock In Kowalski, R., Bowen, K., eds.: Logic Programming: Proc. of the
  Fifth Int'l Conf. and Symp., {MIT} Press (1988)  1070--1080

\bibitem{baral_knowledge_2003}
Baral, C.:
\newblock Knowledge representation, reasoning and declarative problem solving.
\newblock Cambridge University Press (2003)

\bibitem{clark_km:_2004}
Clark, P., Porter, B., Works, B.:
\newblock {KM:} The knowledge machine 2.0: Users manual.
\newblock Citeseer (2004)

\bibitem{herzog2007data}
Herzog, T.N., Scheuren, F.J., Winkler, W.E.:
\newblock Data quality and record linkage techniques.
\newblock Springer (2007)

\bibitem{rahm2000data}
Rahm, E., Do, H.H.:
\newblock Data cleaning: Problems and current approaches.
\newblock IEEE Data Engineering Bulletin \textbf{23}(4) (2000)  3--13

\bibitem{fan2009conditional}
Fan, W., Geerts, F., Jia, X.:
\newblock Conditional dependencies: A principled approach to improving data
  quality.
\newblock In: Dataspace: The Final Frontier.
\newblock Springer (2009)  8--20

\end{thebibliography}
\end{document}